\documentclass[letterpaper, 10pt, conference]{IEEEtran}

\IEEEoverridecommandlockouts
\makeatletter
\def\ps@IEEEtitlepagestyle{%
  \def\@oddfoot{\mycopyrightnotice}%
  \def\@evenfoot{}%
}
\def\mycopyrightnotice{%
  {\hfill \footnotesize This work has been submitted to the IEEE for possible publication. Copyright may be transferred without notice, after which this version may no longer be accessible. \hfill}
}
\makeatother

\usepackage{graphicx}
\usepackage[normalem]{ulem}
\usepackage[flushleft]{threeparttable}

\usepackage{epsfig} 
\usepackage{mathptmx} 
\usepackage{times} 
\usepackage{amsmath} 
\usepackage{amssymb}  
\usepackage{float}
\usepackage{subcaption}
\usepackage{caption}
\usepackage{bm}
\usepackage{romannum}
\usepackage{adjustbox}
\usepackage{textcomp}
\newcommand{\etal}{\textit{et~al}. }
\usepackage[font=small,labelfont=bf]{caption}

\usepackage{hyperref}
\usepackage[dvipsnames]{xcolor}
\newcommand\myshade{85}
\colorlet{mylinkcolor}{violet}
\colorlet{mycitecolor}{YellowOrange}
\colorlet{myurlcolor}{Aquamarine}

\hypersetup{
  linkcolor  = mylinkcolor!\myshade!black,
  citecolor  = mycitecolor!\myshade!black,
  urlcolor   = myurlcolor!\myshade!black,
  colorlinks = true,
}

\usepackage{xcolor}
\usepackage{pifont}
\newcommand{\cmark}{\ding{51}}%
\newcommand{\xmark}{\ding{55}}%

\definecolor{OliveGreen}{rgb}{0,0.5,0.2}
\definecolor{myred}{rgb}{0.8,0.2,0.0}

\begin{document}

\title{\LARGE \bf
Antipodal Robotic Grasping using \\ Generative Residual Convolutional Neural Network
}

\author{Sulabh Kumra, Shirin Joshi and Ferat Sahin
\thanks{Authors are with Multi-Agent Bio-Robotics Laboratory (MABL), Rochester Institute of Technology, Rochester, NY, USA. Sulabh Kumra is also with OSARO Inc., San Francisco, CA, USA.}%
\thanks{Corresponding author e-mail: \href{mailto:sk2881@rit.edu}{sk2881@rit.edu}}
}

\maketitle


\begin{abstract}
In this paper, we present a modular robotic system to tackle the problem of generating and performing antipodal robotic grasps for unknown objects from the n-channel image of the scene. We propose a novel Generative Residual Convolutional Neural Network (GR-ConvNet) model that can generate robust antipodal grasps from n-channel input at real-time speeds ({\raise.17ex\hbox{$\scriptstyle\sim$}}20ms). We evaluate the proposed model architecture on standard datasets and a diverse set of household objects. We achieved state-of-the-art accuracy of 97.7\% and 94.6\% on Cornell and Jacquard grasping datasets, respectively. We also demonstrate a grasp success rate of 95.4\% and 93\% on household and adversarial objects, respectively, using a 7 DoF robotic arm.
\end{abstract}


\section{INTRODUCTION}
Robotic manipulators are constantly compared to humans due to the inherent characteristics of humans to instinctively grasp an unknown object rapidly and with ease based on their own experiences. As increasing research is being done to make the robots more intelligent, there exists a demand for a generalized technique to infer fast and robust grasps for any kind of object that the robot encounters. The major challenge is being able to precisely transfer the knowledge that the robot learns to novel real-world objects.

We present a modular robot agnostic approach to tackle this problem of grasping unknown objects. We propose a Generative Residual Convolutional Neural Network (GR-ConvNet) that generates antipodal grasps for every pixel in an n-channel input image. We use the term generative to distinguish our method from other techniques that output a grasp probability or classify grasp candidates in order to predict the best grasp.

Unlike the previous work done in robotic grasping \cite{lenz2015deep,redmon2015real,pinto2016supersizing,kumra2017robotic}, where the required grasp is predicted as a grasp rectangle calculated by choosing the best grasp from multiple grasp probabilities, our network generates three images from which we can infer grasp rectangles for multiple objects. Additionally, it is possible to infer multiple grasp rectangles for multiple objects from the output of GR-ConvNet in one-shot thereby decreasing the overall computational time.

\begin{figure}
    \centering
    \includegraphics[width=0.48\textwidth]{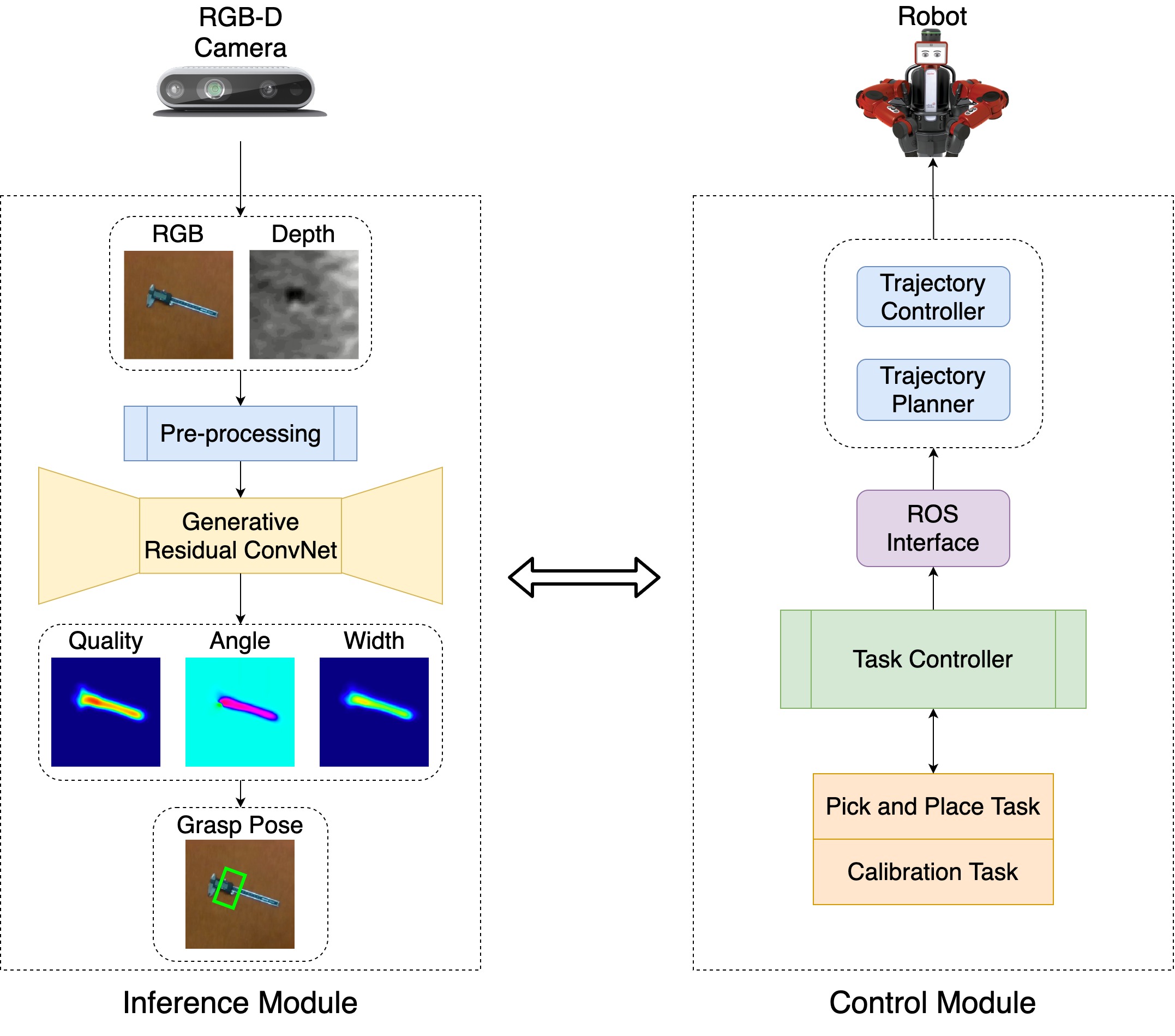}
    \caption{Proposed system overview. Inference module predict suitable grasp poses for the objects in the camera\textquotesingle s field of view. Control module uses these grasp poses to plan and execute robot trajectories to perform antipodal grasps. Video: \href{https://youtu.be/cwlEhdoxY4U}{https://youtu.be/cwlEhdoxY4U}}
    \label{fig: architecture}
\end{figure}

Fig.\ref{fig: architecture} shows an overview of the proposed system architecture. It consists of two main modules: the inference module and the control module. The inference module acquires RGB and aligned depth images of the scene from the RGB-D camera. The images are pre-processed to match the input format of the GR-ConvNet. The network generates quality, angle, and width images, which are then used to infer antipodal grasp poses. The control module consists of a task controller that prepares and executes a plan to perform a pick and place task using the grasp pose generated by the inference module. It communicates the required actions to the robot through a ROS interface using a trajectory planner and controller.


The main contributions of this paper can be summarized as follows: 
\begin{itemize}
    \item We present a modular robotic system that predicts, plans, and performs antipodal grasps for the objects in the scene. We open-sourced the implementation of the proposed inference\footnote{Available at \href{https://github.com/skumra/robotic-grasping}{https://github.com/skumra/robotic-grasping}} and control\footnote{Available at \href{https://github.com/skumra/baxter-pnp}{https://github.com/skumra/baxter-pnp}} modules.
    \item We propose a novel generative residual convolutional neural network architecture that predicts suitable antipodal grasp configurations for objects in the camera\textquotesingle s field of view.
    \item We evaluate our model on publicly available grasping datasets and achieved state-of-the-art accuracy of 97.7\% and 94.6\% on Cornell and Jacquard grasping datasets, respectively.
    \item We demonstrate that the proposed model can be deployed on a robotic arm to perform antipodal grasps at real-time speeds with a success rate of 95.4\% and 93\% on household and adversarial objects, respectively.
\end{itemize}


\section{RELATED WORK}
\textbf{Robotic Grasping}: There has been extensive on-going research in the field of robotics, especially robotic grasping. Although the problem seems to just be able to find a suitable grasp for an object, the actual task involves multifaceted elements such as- the object to be grasped, the shape of the object, physical properties of the object and the gripper with which it needs to be grasped among others. Early research in this field involved hand-engineering the features \cite{maitin2010cloth,kragic2003robust}, which can be a tedious and time-consuming task but can be helpful for learning to grasp objects with multiple fingers such as \cite{kopicki2016one,bohg2013data}. 

Initially for obtaining a stable grasp, the mechanics and contact kinematics of the end effector in contact with the object were studied and the grasp analysis was performed as seen from the survey by \cite{bicchi2000robotic,shimoga1996robot}. Prior work \cite{saxena2008robotic} in robotic grasping for novel objects involved using supervised learning which was trained on synthetic data but it was limited to environments such as office, kitchen, and dishwasher. Satish \etal \cite{satish2019policy} introduced a Fully Convolutional Grasp Quality Convolutional Neural Network (FC-GQ-CNN) which predicted a robust grasp quality by using a data collection policy and synthetic training environment. This method enabled an increase in the number of grasps considered to 5000 times in 0.625s.
However, the current research relies more on using the RGB-D data to predict grasp poses. These approaches depend wholly on deep learning techniques.

\textbf{Deep learning for grasping}: Deep learning has been a hot topic of research since the advent of ImageNet success and the use of GPU\textquotesingle s and other fast computational techniques. Also, the availability of affordable RGB-D sensors enabled the use of deep learning techniques to learn the features of objects directly from image data. Recent experimentations using deep neural networks \cite{redmon2015real,schmidt2018grasping,zeng2018robotic} have demonstrated that they can be used to efficiently compute stable grasps. Pinto \etal \cite{pinto2016supersizing} used an architecture similar to AlexNet which shows that by increasing the size of the data, their CNN was able to generalize better to new data.  Varley \etal \cite{varley2017shape} propose an interesting approach to grasp planning through shape completion where a 3D CNN was used to train the network on the 3D prototype of objects on their own dataset captured from various viewpoints. Guo \etal \cite{guo2017hybrid} used tactile data along with visual data to train a hybrid deep architecture. Mahler \etal \cite{mahler2017dex} proposed a Grasp Quality Convolutional Neural Network (GQ-CNN) that predicts grasps from synthetic point cloud data trained on Dex-Net 2.0 grasp planner dataset. Levine \etal \cite{levine2018learning} discuss the use of monocular images for hand to eye coordination for robotic grasping using a deep learning framework. They use a CNN for grasp success prediction and further use continuous servoing to continuously servo the manipulator to correct mistakes. Antanas \etal \cite{antanas2019semantic} discuss an interesting approach known as a probabilistic logic framework that is said to improve the grasping capability of a robot with the help of semantic object parts. This framework combines high-level reasoning with low-level grasping. The high-level reasoning comprises object affordances, its categories, and task-based information while low-level reasoning uses visual shape features. This has been observed to work well in kitchen-related scenarios.

\renewcommand{\arraystretch}{1.4}
\begin{table}
\begin{center}
\vspace*{0.1cm}

\caption{A comparison of related work}
\label{tab:comparison_related_work}
\begin{adjustbox}{width=.49\textwidth}
\begin{tabular}{lcccccccc}
\hline
 & \cite{lenz2015deep} & \cite{redmon2015real} & \cite{kumra2017robotic} & \cite{morrison2019learning} & \cite{chu2018real} & \cite{zhou2018fully} & \cite{asif2018graspnet} & \textbf{Ours} \\
\hline
\textbf{Real Robot Experiments} & \textcolor{OliveGreen}{\cmark}  & \textcolor{myred}{\xmark} & \textcolor{myred}{\xmark} & \textcolor{OliveGreen}{\cmark} & \textcolor{OliveGreen}{\cmark} & \textcolor{myred}{\xmark} & \textcolor{OliveGreen}{\cmark} & \textcolor{OliveGreen}{\textcolor{OliveGreen}{\cmark}} \\
\textbf{Adversarial Objects}    & \textcolor{myred}{\xmark}  & \textcolor{myred}{\xmark} & \textcolor{myred}{\xmark} & \textcolor{OliveGreen}{\cmark} & \textcolor{myred}{\xmark} & \textcolor{myred}{\xmark} & \textcolor{myred}{\xmark} & \textcolor{OliveGreen}{\cmark} \\
\textbf{Clutter}                & \textcolor{myred}{\xmark}  & \textcolor{myred}{\xmark} & \textcolor{myred}{\xmark} & \textcolor{OliveGreen}{\cmark} & \textcolor{OliveGreen}{\cmark} & \textcolor{myred}{\xmark} & \textcolor{OliveGreen}{\cmark} & \textcolor{OliveGreen}{\cmark} \\
\textbf{Results on Cornell}     & \textcolor{OliveGreen}{\cmark}  & \textcolor{OliveGreen}{\cmark} & \textcolor{OliveGreen}{\cmark} & \textcolor{OliveGreen}{\cmark} & \textcolor{OliveGreen}{\cmark} & \textcolor{OliveGreen}{\cmark} & \textcolor{OliveGreen}{\cmark} & \textcolor{OliveGreen}{\cmark} \\
\textbf{Results on Jacquard}    & \textcolor{myred}{\xmark}  & \textcolor{myred}{\xmark} & \textcolor{myred}{\xmark} & \textcolor{myred}{\xmark} & \textcolor{myred}{\xmark} & \textcolor{OliveGreen}{\cmark} & \textcolor{myred}{\xmark} & \textcolor{OliveGreen}{\cmark} \\
\textbf{Code Available}         & \textcolor{OliveGreen}{\cmark}  & \textcolor{myred}{\xmark} & \textcolor{myred}{\xmark} & \textcolor{OliveGreen}{\cmark} & \textcolor{myred}{\xmark} & \textcolor{myred}{\xmark} & \textcolor{myred}{\xmark} & \textcolor{OliveGreen}{\cmark} \\
\hline
\end{tabular}
\end{adjustbox}
\end{center}
\end{table}

\textbf{Grasping using Uni-modal data} : Johns \etal \cite{johns2016deep} used a simulated depth image to predict a grasp outcome for every grasp pose predicted and select the best grasp by smoothing the predicted pose using a grasp uncertainty function. A generative approach to grasping is discussed by Morrison \etal \cite{morrison2019learning}. The Generative grasp CNN architecture generates grasp poses using a depth image and the network computes grasp on a pixel-wise basis. \cite{morrison2019learning} suggests that it reduces existing shortcomings of discrete sampling and computational complexity. Another recent approach that merely relies on depth data as the sole input to the deep CNN is as seen in \cite{schmidt2018grasping}.

\textbf{Grasping using multi-modal data}: There are different ways of handling objects multi-modalities. Many have used separate features to learn the modalities which can be computationally exhaustive. Wang \etal \cite{wang2016robot} proposed methods that consider multi-modal information as the same. Jiang \etal \cite{jiang2011efficient} used RGB-D images to infer grasps based on a two-step learning process. The first step was used to narrow down the search space and the second step was used to compute the optimal grasp rectangle from the top grasps obtained using the first method. Lenz \etal \cite{lenz2015deep} used a similar two-step approach but with a deep learning architecture which however could not work well on all types of objects and often predicted a grasp location that was not the best grasp for that particular object such as in \cite{jiang2011efficient} the algorithm predicted the grasp for a shoe was from its laces which in practice failed when the robot tried to grasp using the shoelaces while in \cite{lenz2015deep} the algorithm sometimes could not predict grasps which are more practical using just the local information as well as due to the RGB-D sensor used. Yan \etal \cite{yan2019data} used point cloud prediction network to generate a grasp by first preprocessing the data by obtaining the color, depth, and masked image and then obtaining a 3D point cloud of the object to be fed into a critic network to predict a grasp. Chu \etal \cite{chu2018real} propose a novel architecture that can predict multiple grasps for multiple objects simultaneously rather than for a single object. For this, they used a multi-object dataset of their own. The model was also tested on Cornell Grasp Dataset. A robotic grasping method that consists of a ConvNet for object recognition and a grasping method for manipulating the objects is discussed by Ogas \etal \cite{ogas2019robotic}. The grasping method assumes an industry assembly line where the object parameters are assumed to be known in advance. Kumra \etal \cite{kumra2017robotic} proposed a Deep CNN architecture that uses residual layers for predicting robust grasps. The paper demonstrates that a deeper network along with residual layers learns better features and performs faster. Asif \etal \cite{Asif2018EnsembleNetIG} introduced a consolidated framework known as EnsembleNet in which the grasp generation network generates four grasp representations and EnsembleNet synthesizes these generated grasps to produce grasp scores from which the grasp with the highest score gets selected.

Our work is based on similar concepts and is designed to advance the research done in this area. Table \ref{tab:comparison_related_work} provides a comparison of our work to recent related work in robotic grasping for unknown objects.


\section{PROBLEM FORMULATION}
In this work, we define the problem of robotic grasping as predicting antipodal grasps for unknown objects from an n-channel image of the scene and executing it on a robot.

Instead of the 5 dimensional grasp representation used in \cite{lenz2015deep, redmon2015real, kumra2017robotic}, we use an improved version of grasp representation similar to the one proposed by Morrison \etal in \cite{morrison2019learning}. We denote the grasp pose in robot frame as:
\begin{equation}\label{eq:1}
    G_r = (\textbf{P}, \Theta_r, W_r, Q) 
\end{equation}
where, $\textbf{P} = (x, y, z)$ is tool tip\textquotesingle s center position, $\Theta_r$ is tools rotation around the z-axis, $W_r$ is the required width for the tool, and $Q$ is the grasp quality score.

We detect a grasp from an n-channel image $\textbf{I} = \mathbb{R}^{n\times h \times w}$ with height $h$ and width $w$, which can be defined as:
\begin{equation}\label{eq:2}
    G_i = (x, y, \Theta_i, W_i, Q)
\end{equation}
where $(x, y)$ corresponds to the center of grasp in image coordinates, $\Theta_i$ is the rotation in camera\textquotesingle s frame of reference, $W_i$ is the required width in image coordinates, and $Q$ is the same scalar as in equation \eqref{eq:1}.

The grasp quality score $Q$ is the quality of the grasp at every point in the image and is indicated as a score value between 0 and 1 where a value that is in proximity to 1 indicates a greater chance of grasp success. $\Theta_i$ indicates the antipodal measurement of the amount of angular rotation required at each point to grasp the object of interest and is represented as a value in the range $[\frac{-\pi}{2}, \frac{\pi}{2}]$. $W_i$ is the required width which is represented as a measure of uniform depth and indicated as a value in the range of $[0, W_{max}]$ pixels. $W_{max}$ is the maximum width of the antipodal gripper.

To execute a grasp obtained in the image space on a robot, we can apply the following transformations to convert the image coordinates to robot\textquotesingle s frame of reference.
\begin{equation}
    G_r = T_{rc}(T_{ci}(G_i))
\end{equation}
where, $T_{ci}$ is a transformation that converts image space into camera\textquotesingle s 3D space using the intrinsic parameters of the camera, and $T_{rc}$ converts camera space into the robot space using the camera pose calibration value. 

This notation can be scaled for multiple grasps in an image. The collective group of all the grasps can be denoted as:
\begin{equation}
    \textbf{G} = (\bm{\Theta},\textbf{W},\textbf{Q}) \in \mathbb{R}^{3\times h \times w}
\end{equation}
where $\bm{\Theta},\textbf{W}$, and $\textbf{Q}$ represents three images in the form of grasp angle, grasp width and grasp quality score respectively calculated at every pixel of an image using equation \eqref{eq:2}. 


\begin{figure*}
    \centering
    \vspace*{0.2cm}
    \includegraphics[width=1.0\textwidth]{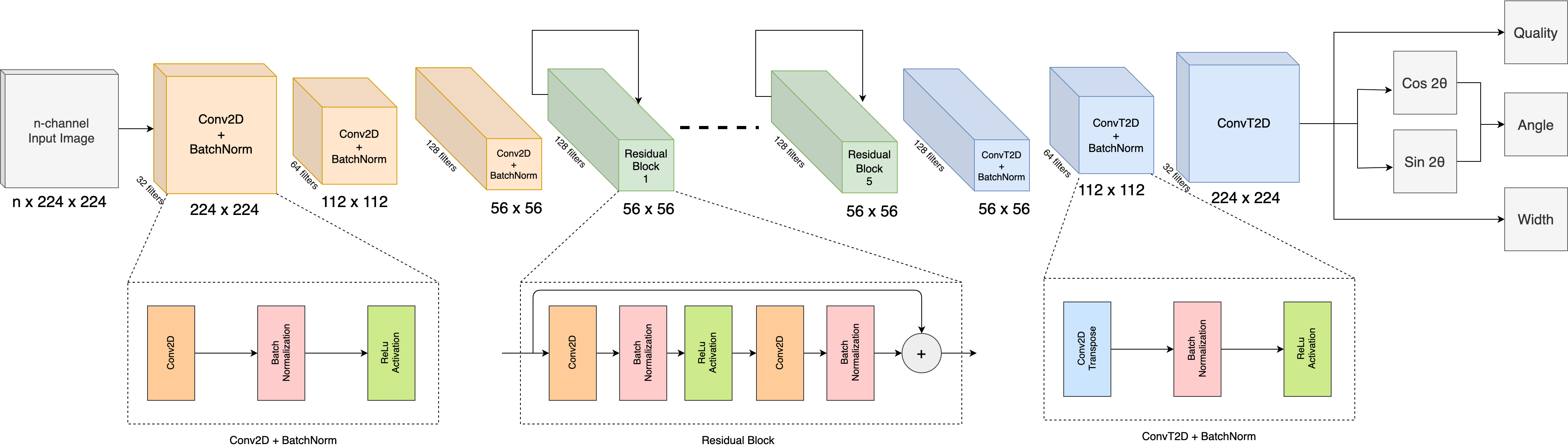}
    \caption{Proposed Generative Residual Convolutional Neural Network}
    \label{fig: network}
\end{figure*}

\section{APPROACH}

We propose a dual-module system to predict, plan and perform antipodal grasps for the objects in the scene. The overview of the proposed system is shown in fig.\ref{fig: architecture}. The inference module is used to predict suitable grasp poses for the objects in the camera\textquotesingle s field of view. The control module uses these grasp poses to plan and execute robot trajectories to perform antipodal grasps.

\subsection{Inference module}
The inference module consists of three parts. First, the input data is pre-processed where it is cropped, resized, and normalized. If the input has a depth image, it is inpainted to obtain a depth representation \cite{xue2017depth}. The $224\times224$ n-channel processed input image is fed into the GR-ConvNet. It uses n-channel input that is not limited to a particular type of input modality such as a depth-only or RGB-only image as our input image. Thus, making it generalized for any kind of input modality. The second generates three images as grasp angle, grasp width, and grasp quality score as the output using the features extracted from the pre-processed image using GR-ConvNet. The third infers grasp poses from the three output images.

\subsection{Control module}
The control module mainly incorporates a task controller that performs tasks such as pick-and-place and calibration. The controller requests a grasp pose from the inference module which returns the grasp pose with the highest quality score. The grasp pose is then converted from camera coordinates into robot coordinates using the transform calculated from hand-eye calibration \cite{strobl2006optimal}. Further, the grasp pose in robot frame is used to plan a trajectory to perform the pick and place action using inverse kinematics through a ROS interface. The robot then executes the planned trajectory. Due to our modular approach and ROS integration, this system can be adapted for any robotic arm.

\subsection{Model architecture}
Fig. \ref{fig: network} shows the proposed GR-ConvNet model, which is a generative architecture that takes in an n-channel input image and generates pixel-wise grasps in the form of three images. The n-channel image is passed through three convolutional layers, followed by five residual layers and convolution transpose layers to generate four images. These output images consist of grasp quality score, required angle in the form of $\cos{2\Theta}$, and $\sin{2\Theta}$ as well as the required width of the end effector. Since the antipodal grasp is uniform around $\pm \frac{\pi}{2}$, we extract the angle in the form of two elements $\cos{2\Theta}$ and $\sin{2\Theta}$ that output distinct values that are combined to form the required angle.

The convolutional layers extract the features from the input image. The output of the convolutional layer is then fed into 5 residual layers. As we know, accuracy increases with increasing the number of layers. However, it is not true when you exceed a certain number of layers, which results in the problem of vanishing gradients and dimensionality error, thereby causing saturation and degradation in the accuracy. Thus, using residual layers enables us to better learn the identity functions by using skip connections. After passing the image through these convolutional and residual layers, the size of the image is reduced to $56\times56$, which can be difficult to interpret. Therefore, to make it easier to interpret and retain spatial features of the image after convolution operation, we up-sample the image by using a convolution transpose operation. Thus, we obtain the same size of the image at the output as the size of the input. 

Our network has a total of 1,900,900 parameters which indicate that our network is comparatively shorter as opposed to other networks \cite{kumra2017robotic, zhou2018fully, Asif2018EnsembleNetIG}. Thereby making it computationally less expensive and faster in contrast to other architectures using similar grasp prediction techniques that contain millions of parameters and complex architectures. The lightweight nature of the model makes it suitable for closed-loop control at a rate of up to 50 Hz.

\subsection{Training methodology}
For a dataset having objects $D = \left \{ D_1 \dots D_n \right \}$, input scene images $I = \left \{ I^1 \dots I^n \right \}$ and successful grasps in image frame $G_i = \left \{ g_1^1 \dots g_{m_1}^1 \dots g_1^2 \dots g_{m_n}^n \right \}$, we can train our model end-to-end to learn the mapping function $\gamma (I , D) = G_i$ by minimizing the negative log-likelihood of $G_i$ conditioned on the input image scene $I$, which is given by:

\begin{equation}
    - \frac{1}{n} \sum_{i=1}^{n} \frac{1}{m_i} \sum_{j=1}^{m_i} log \gamma (g_i^j | I^i)
\end{equation}

The models were trained using the Adam optimizer \cite{adam2015} and standard backpropagation and mini-batch SGD technique \cite{nitanda2014stochastic}. The learning rate was set as ${10^{-3}}$ and a mini-batch size of $8$ was used. We trained the model using three random seeds, and report the average of the three seeds. 

\subsection{Loss function}
We analyzed the performance of various loss functions for our network and after running a few trials found that in order to handle exploding gradients, the smooth L1 loss also known as Huber loss works best. We define our loss as :
\begin{equation}
     \mathcal{L}(G_i, \widehat{G_i}) = \frac{1}{n} \sum_{}^{k} z_k
\end{equation}
where $z_{i}$ is given by:
\begin{equation}
    z_k = \left\{\begin{matrix}
    0.5 (G_{i_k} - \widehat{G_{i_k}})^{2}, & if \left | G_{i_k} - \widehat{G_{i_k}} \right | < 1 \\ 
    \left | G_{i_k} - \widehat{G_{i_k}} \right | - 0.5 & otherwise
    \end{matrix}\right.
\end{equation}
$G_i$ is the grasp generated by the network and $\widehat{G_i}$ is the ground truth grasp.


\section{EVALUATION}

\subsection{Datasets}
There are a limited number of publicly available antipodal grasping datasets. Table \ref{tab:datasets} shows a summary of the publicly available antipodal grasping datasets. We used two of these datasets for training and evaluating our model. The first one is the Cornell grasp dataset \cite{jiang2011efficient}, which is the most common grasping dataset used to benchmark results, and the second one is a more recent Jacquard grasping dataset \cite{depierre2018jacquard}, which is more than 50 times bigger the Cornell grasp dataset.

\begin{table}[htbp]
\vspace{2mm}
\begin{center}
\caption{Summary of Antipodal Grasping Datasets}
\label{tab:datasets}
\begin{tabular}{l|c|c|c|c}
\hline
\textbf{Dataset} & \textbf{Modality} & \textbf{Objects} & \textbf{Images} & \textbf{Grasps} \\
\hline
Cornell & RGB-D & 240 & 1035 & 8019 \\
Dexnet & Depth & 1500 & 6.7M & 6.7M \\
Jacquard & RGB-D & 11k & 54k & 1.1M \\
\hline
\end{tabular}
\end{center}
\end{table}

The extended version of \textbf{Cornell Grasp Dataset} comprises of 1035 RGB-D images with a resolution of $640 \times 480$ pixels of 240 different real objects with 5110 positive and 2909 negative grasps. The annotated ground truth consists of several grasp rectangles representing grasping possibilities per object. However, it is a small dataset for training our GR-ConvNet model, therefore we create an augmented dataset using random crops, zooms, and rotations which effectively has 51k grasp examples. Only positively labeled grasps from the dataset were considered during training. 

The \textbf{Jacquard Grasping Dataset} is built on a subset of ShapeNet which is a large CAD models dataset. It consists of 54k RGB-D images and annotations of successful grasping positions based on grasp attempts performed in a simulated environment. In total, it has 1.1M grasp examples. As this dataset was large enough to train our model, no augmentation was performed.

\subsection{Grasp Detection Metric}
For a fair comparison of our results, we use the rectangle metric \cite{jiang2011efficient} proposed by Jiang \etal to report the performance of our system. According to the proposed rectangle metric, a grasp is considered to be valid when it satisfies the following two conditions:
\begin{itemize}
    \item The intersection over union (IoU) score between the ground truth grasp rectangle and the predicted grasp rectangle is more than $25\%$.
    \item The offset between the grasp orientation of the predicted grasp rectangle and the ground truth rectangle is less than $30^{\circ}$.
\end{itemize}

This metric requires a grasp rectangle representation, but our model predicts image-based grasp representation $\widehat{G_i}$ using equation \ref{eq:2}. Therefore, in order to convert from image-based grasp representation to rectangle representation, the value corresponding to each pixel in the output image is mapped to its equivalent rectangle representation.


\section{EXPERIMENTS}
In our experiments, we evaluate our approach on: (\romannum{1}) two standard datasets, (\romannum{2}) household objects, (\romannum{3}) adversarial objects and (\romannum{4}) objects in clutter.


\begin{figure}
\vspace*{0.1cm}
\begin{subfigure}{.24\textwidth}
  \centering
  \includegraphics[width=1\linewidth]{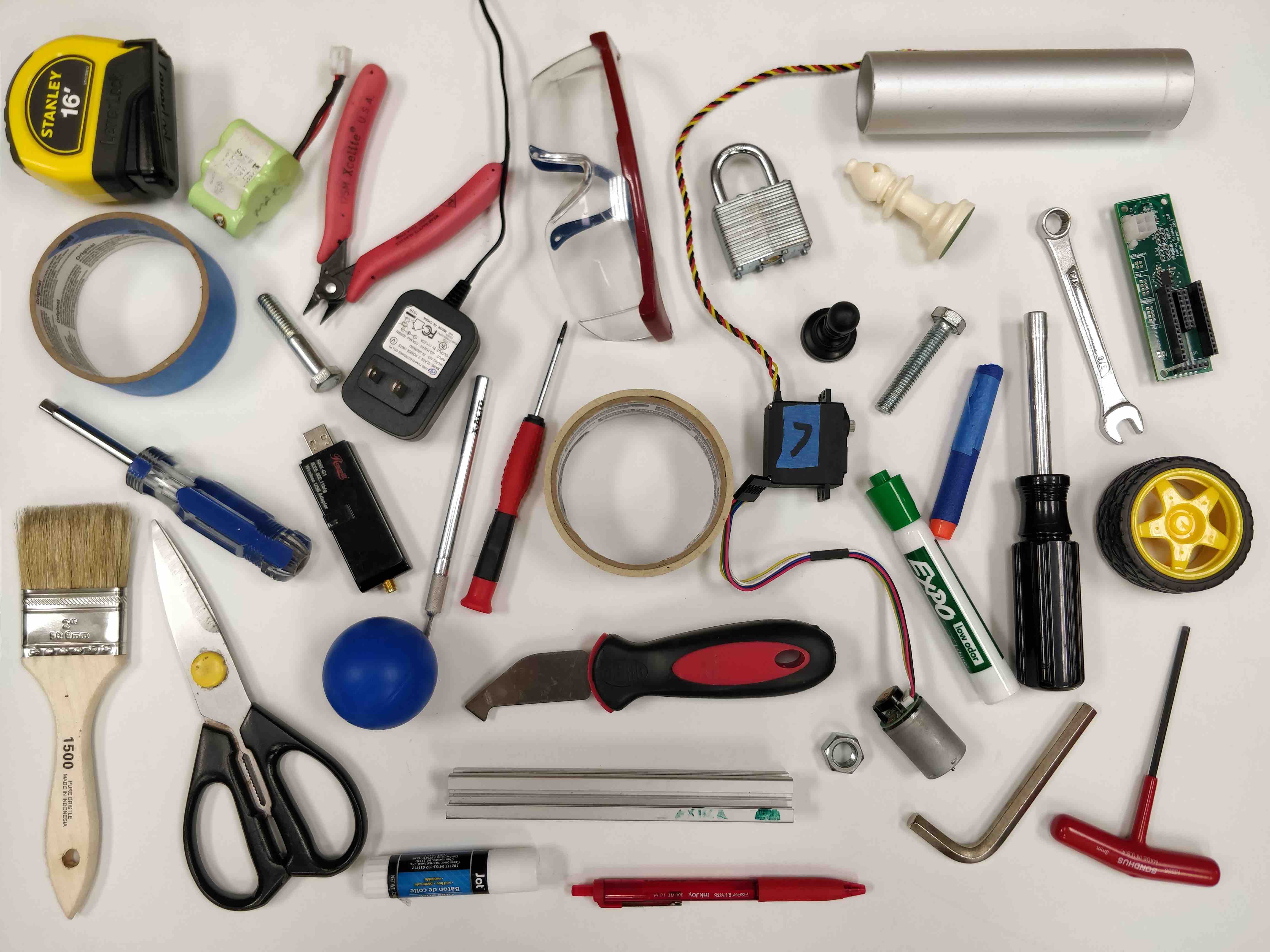}  
  \caption{}
  \label{fig:household_objects}
\end{subfigure}
\begin{subfigure}{.24\textwidth}
  \centering
  \includegraphics[width=1\linewidth]{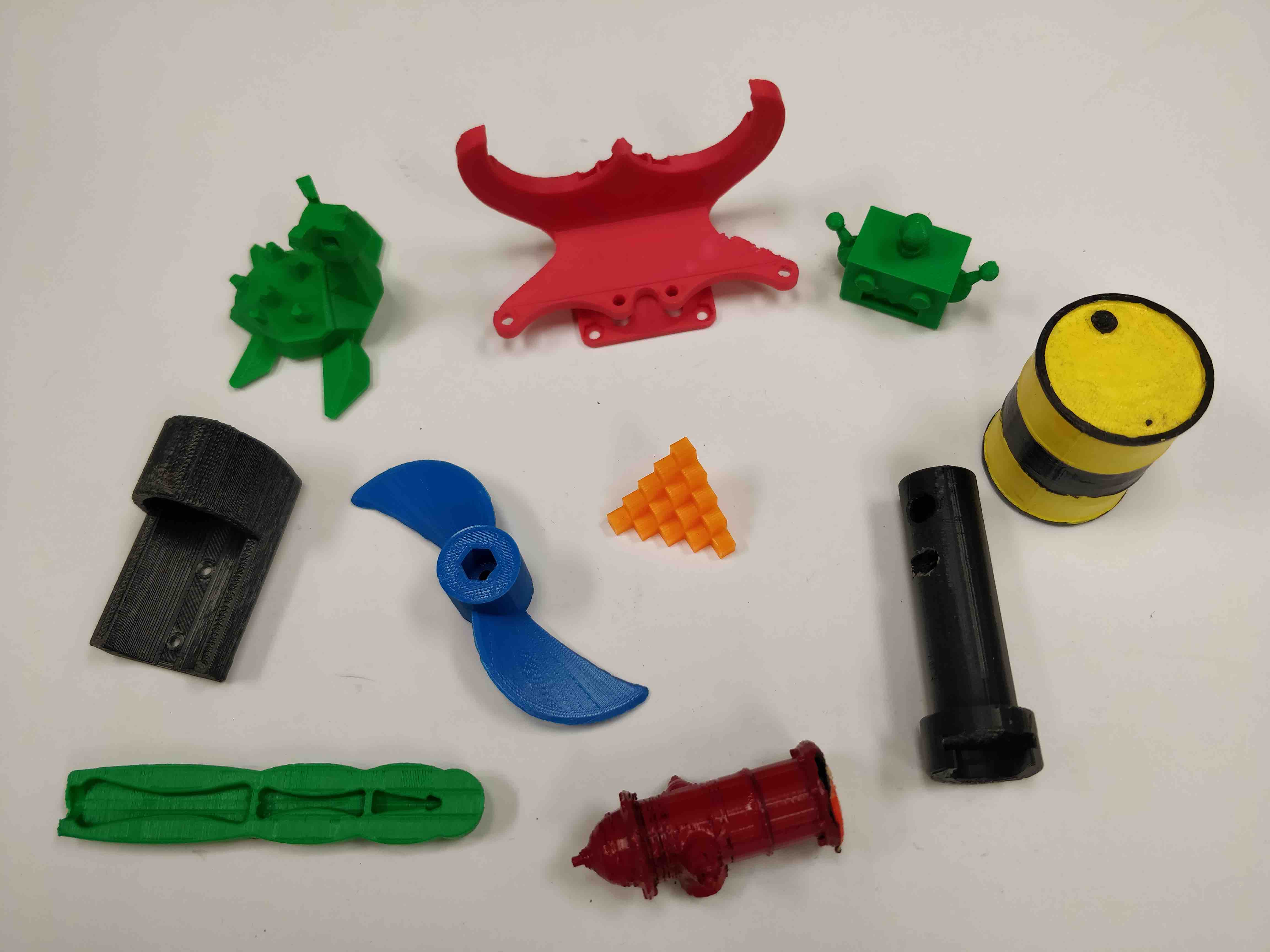}  
  \caption{}
  \label{fig:adversarial_objects}
\end{subfigure}
\caption{Objects used for robotic grasping experiments. (a) Household test objects. (b) Adversarial test objects.}
\label{fig:test_objects}
\end{figure}

\subsection{Setup}
To get the scene image for the real-world experiments, we used the Intel RealSense Depth Camera D435 that uses stereo vision to calculate depth. It consists of a pair of RGB sensors, depth sensors, and an infrared projector. The experiments were conducted on the 7-DoF Baxter Robot by Rethink Robotics. A two-fingered parallel gripper was used for grasping the test objects. The camera was mounted behind the robot arm looking over the shoulder.

The execution times for our proposed GR-ConvNet are measured on a system running Ubuntu 16.04 with an Intel Core i7-7800X CPU clocked at 3.50 GHz and an NVIDIA GeForce GTX 1080 Ti graphics card with CUDA 10.

\subsection{Household test objects}
A total of 35 household objects were chosen for testing the performance of our system. Each object was tested individually for 10 different positions and orientations which resulted in 350 grasp attempts. The objects were chosen such that each object represented different shape, size, and geometry; and had minimum or no resemblance with each other. We created a mix of deformable, difficult to grasp, reflective, and small objects that need high precision. Fig. \ref{fig:household_objects} shows the set of objects that were used for the experiments.

\subsection{Adversarial test objects}
Another set consisting of 10 adversarial objects with complex geometry was used to evaluate the accuracy of our proposed system. These 3D printed objects have abstract geometry with indefinite surfaces and edges that are hard to perceive and grasp. Each of these objects was tested in isolation for 10 different orientations and positions and made up of a total of 100 grasp attempts. Fig. \ref{fig:adversarial_objects} shows the adversarial objects used during the experiments.

\subsection{Objects in clutter}
Industrial applications such as warehouses require objects to be picked in isolation as well as from a clutter. Therefore, to perform our experiments on cluttered objects we carried out 10 runs with 60 unseen objects. A set of distinct objects for each run was selected from the previously unseen novel objects to create a cluttered scene. An example of this is shown in fig. \ref{fig: example_run}. Each run is terminated when there are no objects in the camera\textquotesingle s field of view. 


\begin{figure*}
    \centering
    \includegraphics[width=1\linewidth]{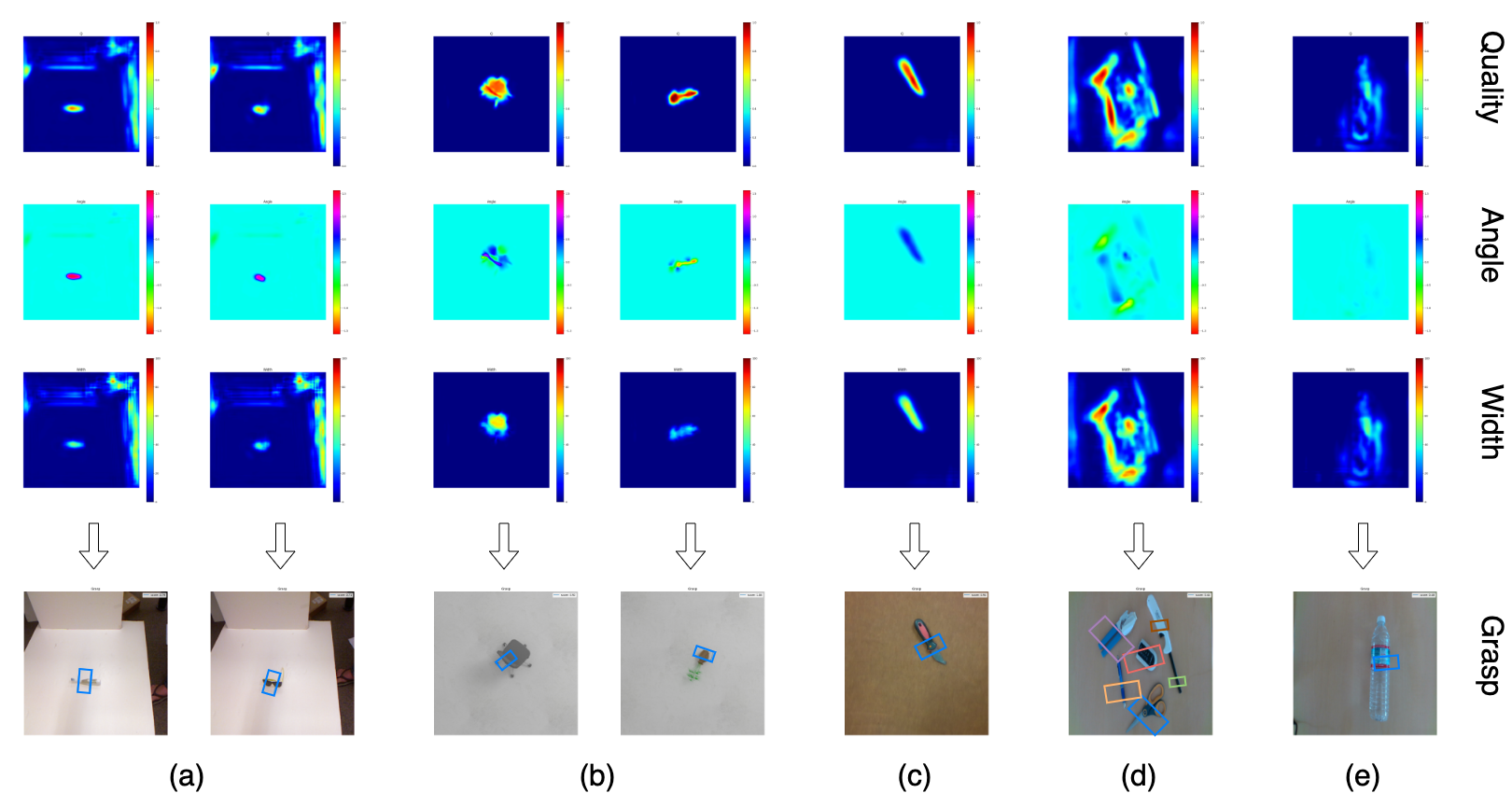}  
    \caption{Qualitative results. Quality, angle and width are the output of GR-ConNet which are used to infer grasp rectangle. (a) Unseen objects from Cornell dataset (b) Unseen objects from Jacquard dataset (c) Single household object (d) Multiple grasps for multiple objects (e) Poor grasp for transparent object}
    \label{fig:all_results}
\end{figure*}

\section{RESULTS} \label{results}
In this section, we discuss the results of our experiments. We evaluate GR-ConvNet on both the Cornell and the Jacquard dataset to examine the outcomes for each of the datasets based on factors such as the size of the dataset, type of training data and demonstrate our model\textquotesingle s capacity to generalize to any kind of object. Further, we show that our model is able to not only generate a single grasp for isolated objects but also multiple grasps for multiple objects in clutter.

Fig. \ref{fig:all_results} shows the qualitative results obtained on previously unseen objects. The figure consists of output in image representation $G_i$ in the form of grasp quality score $Q$, the required angle for grasping $\Theta_i$, and the required gripper width $W_i$. It also includes the output in the form of a rectangle grasp representation projected on the RGB image.


Further, we demonstrate the viability of our method in comparison to other methods by gauging the performance of our network on different types of objects. Additionally, we evaluate the performance of our network on different input modalities. The modalities that the model was tested on included uni-modal input such as depth only and RGB only input images; and multi-modal input such as RGB-D images. Table \ref{tab:cornell_results} show that our network performed better on multi-modal data as compared to uni-modal data since multiple input modalities enabled better learning of the input features.

\begin{table}
\begin{center}
\vspace*{0.1cm}
\caption{Results on the Cornell Dataset}
\label{tab:cornell_results}
\begin{tabular}{l|l|cc|c}
\hline
\textbf{Authors} & \textbf{Algorithm} & \multicolumn{2}{|c|}{\textbf{Accuracy (\%)}} & \textbf{Speed} \\
 & & IW & OW & (ms)\\
\hline
Jiang \cite{jiang2011efficient} & Fast Search & 60.5 & 58.3 & 5000 \\
Lenz \cite{lenz2015deep} & SAE, struct. reg. & 73.9 & 75.6 & 1350 \\
Redmon \cite{redmon2015real} & AlexNet, MultiGrasp & 88.0 & 87.1 & 76 \\
Wang \cite{wang2016robot} & Two-stage closed-loop & 85.3 & - & 140 \\
Asif \cite{asif2017rgb} & STEM-CaRFs & 88.2 & 87.5 & - \\
Kumra \cite{kumra2017robotic} & ResNet-50x2 & 89.2 & 88.9 & 103 \\
Morrison \cite{morrison2019learning} & GG-CNN & 73.0 & 69.0 & 19 \\
Guo \cite{guo2017hybrid} & ZF-net & 93.2 & 89.1 & - \\
Zhou \cite{zhou2018fully} & FCGN, ResNet-101 & 97.7 & 96.6 & 117 \\
Karaoguz \cite{karaoguz2019object} & GRPN & 88.7 & - & 200 \\
Asif \cite{asif2018graspnet} & GraspNet & 90.2 & 90.6 & 24 \\
\hline
 & GR-ConvNet-D & 93.2 & 94.3 & 19 \\
Our & GR-ConvNet-RGB & 96.6 & 95.5 & 19 \\
 & GR-ConvNet-RGB-D & \textbf{97.7} & \textbf{96.6} & \textbf{20} \\
\hline
\end{tabular}
\end{center}
\end{table}

\subsection{Cornell Dataset} \label{Cornell results}
We follow a cross-validation setup as in previous works \cite{lenz2015deep, redmon2015real, kumra2017robotic, asif2018graspnet, guo2017hybrid}, using image-wise (IW) and object-wise (OW) data splits. Table \ref{tab:cornell_results} shows the performance of our system for multiple modalities in comparison to other techniques used for grasp prediction. We obtained state-of-the-art accuracy of 97.7\% on Image-wise split and 96.6\% on Object-wise split using RGB-D data, outperforming all competitive methods as seen in table \ref{tab:cornell_results}. The results obtained on the previously unseen objects in the dataset depict that our network can predict robust grasps for different types of objects in the validation set. The data augmentation performed on the Cornell grasp dataset improved the overall performance of the network. Further, the recorded prediction speed of 20ms per image suggests that GR-ConvNet is suitable for real-time closed-loop applications.

\subsection{Jacquard Dataset} \label{Jacquard results}
For the Jacquard dataset, we trained our network on 90\% of the dataset images and validated on 10\% of the remaining dataset. As the Jacquard dataset is much larger than the Cornell dataset, no data augmentation was required. We performed experiments on the Jacquard dataset using multiple modalities and obtained state-of-the-art results with an accuracy of 94.6\% using RGB-D data as the input. Table \ref{tab:jacquard_results} shows that our network not only gives the best results on the Cornell grasp dataset but also outperforms other methods on the Jacquard dataset.

\begin{table}
\vspace*{0.1cm}
\begin{center}
\caption{Results on the Jacquard Dataset}
\label{tab:jacquard_results}
\begin{tabular}{llc}
\hline
\textbf{Authors} & \textbf{Algorithm} & \textbf{Accuracy (\%)} \\
\hline
Depierre \cite{depierre2018jacquard} & Jacquard & 74.2 \\
Morrison \cite{morrison2019learning} & GG-CNN2 & 84 \\
Zhou \cite{zhou2018fully} & FCGN, ResNet-101 & 91.8 \\
\hline
 & GR-ConvNet - D & 93.7 \\
Our & GR-ConvNet - RGB & 91.8 \\
 & GR-ConvNet - RGB-D & \textbf{94.6} \\
\hline
\end{tabular}
\end{center}
\end{table}

\subsection{Grasping novel objects}
Along with the state-of-the-art results on two standard datasets, we also demonstrate that our system equally outperforms in robotic grasping experiments for novel real-world objects. We used 35 household and 10 adversarial objects to evaluate the performance of our system in the physical world using the Baxter robotic arm. Each of the objects was tested for 10 different positions and orientations. The robot performed 334 successful grasps of the total 350 grasp attempts on household objects resulting in an accuracy of 95.4\% and 93 successful grasps out of 100 grasp attempts on adversarial objects giving an accuracy of 93\%. Table \ref{tab:novel_results} shows our results in comparison to other deep learning based approaches in robotic grasping. 

The results obtained in table \ref{tab:novel_results} and fig. \ref{fig:all_results} indicates that GR-ConvNet is able to generalize well to new objects that it has never seen before. The model was able to generate grasps for all the objects except for a transparent bottle.

\begin{figure}
    \centering
    \includegraphics[width=0.49\textwidth]{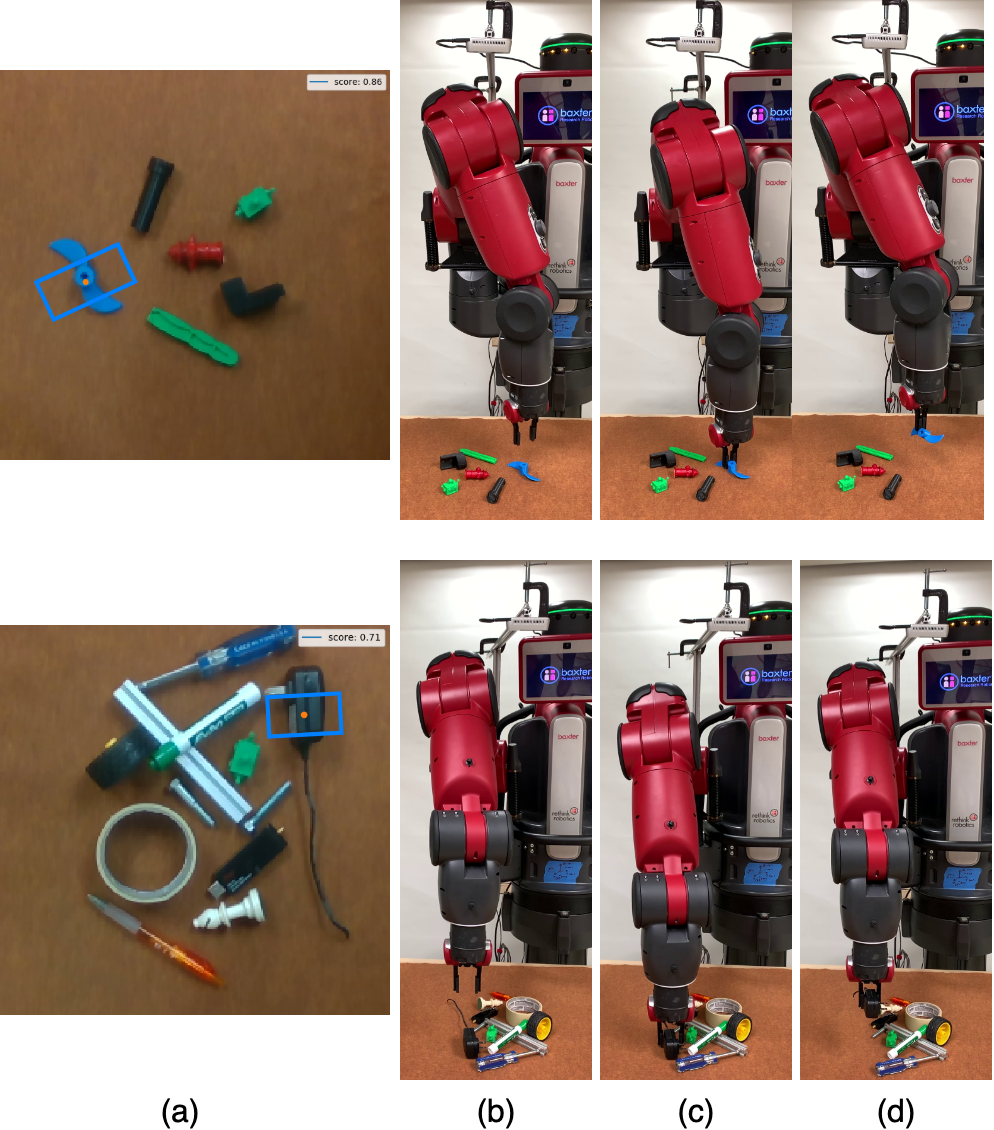}
    \caption{Example of robot grasping adversarial (top) and household (bottom) objects. See attached video for complete run. (a) Grasp pose generated by inference module. (b) Robot approaching the grasp pose. (c) Robot grasping the object. (d) Robot retracting after successful grasp.}
    \label{fig: example_run}
\end{figure}

\subsection{Objects in clutter}
Along with predicting optimum grasps for novel real objects, our robust model is able to predict multiple antipodal grasps for multiple objects in clutter. Each run was performed with as well as without object replacement, and we achieved a grasp success of 93.5\% by averaging grasp success for every successful grasp attempt in each run. Despite the model being trained only on isolated objects, it was able to efficiently predict grasps for manifold objects. Moreover, fig. \ref{fig:all_results}(d) shows grasps predicted for multiple objects and fig. \ref{fig: example_run} illustrates robot grasping household and adversarial objects in cluttered environments. This demonstrates that GR-ConvNet generalizes to all types of objects and can predict robust grasps for multiple objects in clutter. 

\subsection{Failure case analysis}
In our experimental results, there are only a few cases that can be accounted for as failures. Of them, the objects that had extremely low grasp scores and those that slipped from the gripper in spite of the gripper being closed were the most common ones. This could be attributed to the inaccurate depth information coming from the camera and the gripper misalignment due to collision between the gripper and nearby objects.

Another case where the model was unable to produce a good grasp was for the transparent bottle as seen in fig. \ref{fig:all_results}(e). This could be due to inaccurate depth data captured by the camera because of possible object reflections. However, by combining depth data along with RGB data, the model was still able to generate a fairly good grasp for the transparent objects.

\begin{table}
\begin{center}
\caption{Results from robotic grasping experiments}
\label{tab:novel_results}
\begin{tabular}{l c c}
\hline
\textbf{Approach} & \textbf{Household Objects} & \textbf{Adversarial Objects} \\
 & Accuracy (\%) & Accuracy (\%) \\
\hline
\cite{lenz2015deep} & 89 (89/100) & - \\
\cite{pinto2016supersizing} & 73 (109/150) & - \\
\cite{morrison2019learning} & 92 (110/120) & 84 (67/80) \\
\cite{chu2018real} & 89 (89/100) & - \\
\hline
\textbf{Ours} & \textbf{95.4} (334/350) & \textbf{93} (93/100) \\
\hline
\end{tabular}
\end{center}
\end{table}


\section{CONCLUSION}
We presented a modular solution for grasping novel objects using our Generative Residual Convolutional Neural Network that uses n-channel input data to generate images that can be used to infer grasp rectangles for each pixel in an image. We evaluated the GR-ConvNet on two standard datasets, the Cornell grasp dataset and the Jacquard dataset and obtained state-of-the-art results on both the datasets. We also validated the proposed system on novel real objects in clutter using a robotic arm. The results demonstrate that our system can predict and perform accurate grasps for previously unseen objects. Moreover, the low inference time of our model makes the system suitable for closed-loop robotic grasping.

In future work, we would like to extend our solution for different types of grippers used such as single and multiple suction cups and multi-fingered grippers. We would also like to use depth prediction techniques to accurately predict depth for reflective objects, which can aid in improving the grasp prediction accuracy for reflective objects like the bottle.


\section*{Acknowledgment}

The authors acknowledge Research Computing \cite{ritrc} at the Rochester Institute of Technology for providing computational resources and support that have contributed to the research results reported in this publication.


\bibliographystyle{IEEEtran} 
\bibliography{references} 

\begin{thebibliography}{10}
\providecommand{\url}[1]{#1}
\csname url@rmstyle\endcsname
\providecommand{\newblock}{\relax}
\providecommand{\bibinfo}[2]{#2}
\providecommand\BIBentrySTDinterwordspacing{\spaceskip=0pt\relax}
\providecommand\BIBentryALTinterwordstretchfactor{4}
\providecommand\BIBentryALTinterwordspacing{\spaceskip=\fontdimen2\font plus
\BIBentryALTinterwordstretchfactor\fontdimen3\font minus
  \fontdimen4\font\relax}
\providecommand\BIBforeignlanguage[2]{{%
\expandafter\ifx\csname l@#1\endcsname\relax
\typeout{** WARNING: IEEEtran.bst: No hyphenation pattern has been}%
\typeout{** loaded for the language `#1'. Using the pattern for}%
\typeout{** the default language instead.}%
\else
\language=\csname l@#1\endcsname
\fi
#2}}

\bibitem{lenz2015deep}
I.~Lenz, H.~Lee, and A.~Saxena, ``Deep learning for detecting robotic grasps,''
  \emph{The International Journal of Robotics Research}, vol.~34, no. 4-5, pp.
  705--724, 2015.

\bibitem{redmon2015real}
J.~Redmon and A.~Angelova, ``Real-time grasp detection using convolutional
  neural networks,'' in \emph{2015 IEEE International Conference on Robotics
  and Automation (ICRA)}.\hskip 1em plus 0.5em minus 0.4em\relax IEEE, 2015,
  pp. 1316--1322.

\bibitem{pinto2016supersizing}
L.~Pinto and A.~Gupta, ``Supersizing self-supervision: Learning to grasp from
  50k tries and 700 robot hours,'' in \emph{2016 IEEE international conference
  on robotics and automation (ICRA)}.\hskip 1em plus 0.5em minus 0.4em\relax
  IEEE, 2016, pp. 3406--3413.

\bibitem{kumra2017robotic}
S.~Kumra and C.~Kanan, ``Robotic grasp detection using deep convolutional
  neural networks,'' in \emph{2017 IEEE/RSJ International Conference on
  Intelligent Robots and Systems (IROS)}.\hskip 1em plus 0.5em minus
  0.4em\relax IEEE, 2017, pp. 769--776.

\bibitem{maitin2010cloth}
J.~Maitin-Shepard, M.~Cusumano-Towner, J.~Lei, and P.~Abbeel, ``Cloth grasp
  point detection based on multiple-view geometric cues with application to
  robotic towel folding,'' in \emph{2010 IEEE International Conference on
  Robotics and Automation}.\hskip 1em plus 0.5em minus 0.4em\relax IEEE, 2010,
  pp. 2308--2315.

\bibitem{kragic2003robust}
D.~Kragic and H.~I. Christensen, ``Robust visual servoing,'' \emph{The
  International Journal of Robotics Research}, vol.~22, no. 10-11, pp.
  923--939, 2003.

\bibitem{kopicki2016one}
M.~Kopicki, R.~Detry, M.~Adjigble, R.~Stolkin, A.~Leonardis, and J.~L. Wyatt,
  ``One-shot learning and generation of dexterous grasps for novel objects,''
  \emph{The International Journal of Robotics Research}, vol.~35, no.~8, pp.
  959--976, 2016.

\bibitem{bohg2013data}
J.~Bohg, A.~Morales, T.~Asfour, and D.~Kragic, ``Data-driven grasp
  synthesis—a survey,'' \emph{IEEE Transactions on Robotics}, vol.~30, no.~2,
  pp. 289--309, 2013.

\bibitem{bicchi2000robotic}
A.~Bicchi and V.~Kumar, ``Robotic grasping and contact: A review,'' in
  \emph{Proceedings 2000 ICRA. Millennium Conference. IEEE International
  Conference on Robotics and Automation. Symposia Proceedings (Cat. No.
  00CH37065)}, vol.~1.\hskip 1em plus 0.5em minus 0.4em\relax IEEE, 2000, pp.
  348--353.

\bibitem{shimoga1996robot}
K.~B. Shimoga, ``Robot grasp synthesis algorithms: A survey,'' \emph{The
  International Journal of Robotics Research}, vol.~15, no.~3, pp. 230--266,
  1996.

\bibitem{saxena2008robotic}
A.~Saxena, J.~Driemeyer, and A.~Y. Ng, ``Robotic grasping of novel objects
  using vision,'' \emph{The International Journal of Robotics Research},
  vol.~27, no.~2, pp. 157--173, 2008.

\bibitem{satish2019policy}
V.~Satish, J.~Mahler, and K.~Goldberg, ``On-policy dataset synthesis for
  learning robot grasping policies using fully convolutional deep networks,''
  \emph{IEEE Robotics and Automation Letters}, vol.~4, no.~2, pp. 1357--1364,
  2019.

\bibitem{schmidt2018grasping}
P.~Schmidt, N.~Vahrenkamp, M.~W{\"a}chter, and T.~Asfour, ``Grasping of unknown
  objects using deep convolutional neural networks based on depth images,'' in
  \emph{2018 IEEE International Conference on Robotics and Automation
  (ICRA)}.\hskip 1em plus 0.5em minus 0.4em\relax IEEE, 2018, pp. 6831--6838.

\bibitem{zeng2018robotic}
A.~Zeng, S.~Song, K.-T. Yu, E.~Donlon, F.~R. Hogan, M.~Bauza, D.~Ma, O.~Taylor,
  M.~Liu, E.~Romo, \emph{et~al.}, ``Robotic pick-and-place of novel objects in
  clutter with multi-affordance grasping and cross-domain image matching,'' in
  \emph{2018 IEEE International Conference on Robotics and Automation
  (ICRA)}.\hskip 1em plus 0.5em minus 0.4em\relax IEEE, 2018, pp. 1--8.

\bibitem{varley2017shape}
J.~Varley, C.~DeChant, A.~Richardson, J.~Ruales, and P.~Allen, ``Shape
  completion enabled robotic grasping,'' in \emph{2017 IEEE/RSJ International
  Conference on Intelligent Robots and Systems (IROS)}.\hskip 1em plus 0.5em
  minus 0.4em\relax IEEE, 2017, pp. 2442--2447.

\bibitem{guo2017hybrid}
D.~Guo, F.~Sun, H.~Liu, T.~Kong, B.~Fang, and N.~Xi, ``A hybrid deep
  architecture for robotic grasp detection,'' in \emph{2017 IEEE International
  Conference on Robotics and Automation (ICRA)}.\hskip 1em plus 0.5em minus
  0.4em\relax IEEE, 2017, pp. 1609--1614.

\bibitem{mahler2017dex}
J.~Mahler, J.~Liang, S.~Niyaz, M.~Laskey, R.~Doan, X.~Liu, J.~A. Ojea, and
  K.~Goldberg, ``Dex-net 2.0: Deep learning to plan robust grasps with
  synthetic point clouds and analytic grasp metrics,'' \emph{arXiv preprint
  arXiv:1703.09312}, 2017.

\bibitem{levine2018learning}
S.~Levine, P.~Pastor, A.~Krizhevsky, J.~Ibarz, and D.~Quillen, ``Learning
  hand-eye coordination for robotic grasping with deep learning and large-scale
  data collection,'' \emph{The International Journal of Robotics Research},
  vol.~37, no. 4-5, pp. 421--436, 2018.

\bibitem{antanas2019semantic}
L.~Antanas, P.~Moreno, M.~Neumann, R.~P. de~Figueiredo, K.~Kersting,
  J.~Santos-Victor, and L.~De~Raedt, ``Semantic and geometric reasoning for
  robotic grasping: a probabilistic logic approach,'' \emph{Autonomous Robots},
  vol.~43, no.~6, pp. 1393--1418, 2019.

\bibitem{morrison2019learning}
D.~Morrison, P.~Corke, and J.~Leitner, ``Learning robust, real-time, reactive
  robotic grasping,'' \emph{The International Journal of Robotics Research}, p.
  0278364919859066, 2019.

\bibitem{chu2018real}
F.-J. Chu, R.~Xu, and P.~A. Vela, ``Real-world multiobject, multigrasp
  detection,'' \emph{IEEE Robotics and Automation Letters}, vol.~3, no.~4, pp.
  3355--3362, 2018.

\bibitem{zhou2018fully}
X.~Zhou, X.~Lan, H.~Zhang, Z.~Tian, Y.~Zhang, and N.~Zheng, ``Fully
  convolutional grasp detection network with oriented anchor box,'' in
  \emph{2018 IEEE/RSJ International Conference on Intelligent Robots and
  Systems (IROS)}.\hskip 1em plus 0.5em minus 0.4em\relax IEEE, 2018, pp.
  7223--7230.

\bibitem{asif2018graspnet}
U.~Asif, J.~Tang, and S.~Harrer, ``Graspnet: An efficient convolutional neural
  network for real-time grasp detection for low-powered devices.'' in
  \emph{IJCAI}, 2018, pp. 4875--4882.

\bibitem{johns2016deep}
E.~Johns, S.~Leutenegger, and A.~J. Davison, ``Deep learning a grasp function
  for grasping under gripper pose uncertainty,'' in \emph{2016 IEEE/RSJ
  International Conference on Intelligent Robots and Systems (IROS)}.\hskip 1em
  plus 0.5em minus 0.4em\relax IEEE, 2016, pp. 4461--4468.

\bibitem{wang2016robot}
Z.~Wang, Z.~Li, B.~Wang, and H.~Liu, ``Robot grasp detection using multimodal
  deep convolutional neural networks,'' \emph{Advances in Mechanical
  Engineering}, vol.~8, no.~9, p. 1687814016668077, 2016.

\bibitem{jiang2011efficient}
Y.~Jiang, S.~Moseson, and A.~Saxena, ``Efficient grasping from rgbd images:
  Learning using a new rectangle representation,'' in \emph{2011 IEEE
  International Conference on Robotics and Automation}.\hskip 1em plus 0.5em
  minus 0.4em\relax IEEE, 2011, pp. 3304--3311.

\bibitem{yan2019data}
X.~Yan, M.~Khansari, J.~Hsu, Y.~Gong, Y.~Bai, S.~Pirk, and H.~Lee,
  ``Data-efficient learning for sim-to-real robotic grasping using deep point
  cloud prediction networks,'' \emph{arXiv preprint arXiv:1906.08989}, 2019.

\bibitem{ogas2019robotic}
E.~Ogas, L.~Avila, G.~Larregay, and D.~Moran, ``A robotic grasping method using
  convnets,'' in \emph{2019 Argentine Conference on Electronics (CAE)}.\hskip
  1em plus 0.5em minus 0.4em\relax IEEE, 2019, pp. 21--26.

\bibitem{Asif2018EnsembleNetIG}
U.~Asif, J.~Tang, and S.~Harrer, ``Ensemblenet: Improving grasp detection using
  an ensemble of convolutional neural networks,'' in \emph{BMVC}, 2018.

\bibitem{xue2017depth}
H.~Xue, S.~Zhang, and D.~Cai, ``Depth image inpainting: Improving low rank
  matrix completion with low gradient regularization,'' \emph{IEEE Transactions
  on Image Processing}, vol.~26, no.~9, pp. 4311--4320, 2017.

\bibitem{strobl2006optimal}
K.~H. Strobl and G.~Hirzinger, ``Optimal hand-eye calibration,'' in \emph{2006
  IEEE/RSJ international conference on intelligent robots and systems}.\hskip
  1em plus 0.5em minus 0.4em\relax IEEE, 2006, pp. 4647--4653.

\bibitem{adam2015}
J.~B. Diederik P.~Kingma, ``Adam: A method for stochastic optimization,''
  \emph{International Conference for Learning Representations}, 2015.

\bibitem{nitanda2014stochastic}
A.~Nitanda, ``Stochastic proximal gradient descent with acceleration
  techniques,'' in \emph{Advances in Neural Information Processing Systems},
  2014, pp. 1574--1582.

\bibitem{depierre2018jacquard}
A.~Depierre, E.~Dellandr{\'e}a, and L.~Chen, ``Jacquard: A large scale dataset
  for robotic grasp detection,'' in \emph{IEEE/RSJ International Conference on
  Intelligent Robots and Systems}, 2018.

\bibitem{asif2017rgb}
U.~Asif, M.~Bennamoun, and F.~A. Sohel, ``Rgb-d object recognition and grasp
  detection using hierarchical cascaded forests,'' \emph{IEEE Transactions on
  Robotics}, 2017.

\bibitem{karaoguz2019object}
H.~Karaoguz and P.~Jensfelt, ``Object detection approach for robot grasp
  detection,'' in \emph{2019 International Conference on Robotics and
  Automation (ICRA)}.\hskip 1em plus 0.5em minus 0.4em\relax IEEE, 2019, pp.
  4953--4959.

\bibitem{ritrc}
\BIBentryALTinterwordspacing
{Rochester Institute of Technology}, ``Research computing services,'' 2019.
  [Online]. Available: \url{https://www.rit.edu/researchcomputing/}
\BIBentrySTDinterwordspacing

\end{thebibliography}

\end{document}